\documentclass[10pt,logo,copyright]{giga-report}
\usepackage[authoryear,sort&compress,round]{natbib}

\usepackage[utf8]{inputenc} %
\usepackage[T1]{fontenc}    %

\usepackage{parskip}        %
\usepackage{url}            %
\usepackage{booktabs}       %
\usepackage{amsfonts}       %
\usepackage{nicefrac}       %
\usepackage{microtype}      %
\usepackage{xcolor}         %
\usepackage[dvipsnames]{xcolor} %
\usepackage{graphicx}
\usepackage{algorithmic}
\usepackage{algorithm}
\usepackage{animate}        %
\usepackage{subcaption}
\usepackage{tabularx}
\usepackage{makecell}
\usepackage{adjustbox}
\usepackage{setspace}
\usepackage{todonotes}
\usepackage{colortbl}
\usepackage{wrapfig}

\newcolumntype{M}[1]{>{\centering\arraybackslash}m{#1}}
\usepackage{float}
\usepackage{tikz}
\usetikzlibrary{positioning,shapes,arrows}
\usepackage{amsmath,amsfonts,bm, bbm,leftindex}
\usepackage{multirow}
\usepackage{comment}
\usepackage{gensymb}
\usepackage{lipsum}
\usetikzlibrary{arrows.meta, positioning, fit}
\usepackage[para]{threeparttable}
\usepackage{tikz}
\usetikzlibrary{tikzmark}

\usepackage{amsmath} 
\usepackage{graphicx}
\usepackage{marvosym}
\usepackage[nameinlink]{cleveref}
\crefname{equation}{Eq.}{Eqs.}
\crefname{figure}{Fig.}{Figs.}
\crefname{section}{Sec.}{Sec.}
\crefname{appendix}{App.}{App.}
\crefname{table}{Tab.}{Tabs.}
\crefname{algorithm}{Algo}{Algo}
\crefname{thm}{Thm}{Thm}
\Crefname{thm}{Thm}{Thm}
\crefname{prop}{Prop}{Prop}
\usepackage[justification=justified,singlelinecheck=false]{caption}
\usepackage{booktabs}
\usepackage{multirow}
\newcommand{\crefnames}[3]{
  \@for\next:=#1\do{
    \expandafter\crefname\expandafter{\next}{#2}{#3}%
  }
}

\title{\centering Zero2Skill: Bootstrapping Robot Skills through Autonomous Data Collection, Training, and Deployment}
\newcommand{\method}{Zero2Skill}

\author{
\vspace{-0.1in}

\footnotesize

\normalfont
\centerline{
Boyuan Wang\textsuperscript{\rm 1,\rm 2,*}, Zhenyuan Zhang\textsuperscript{\rm 1,\rm 3,*}, Zhiqin Yang\textsuperscript{\rm 3,*}, Peijun Gu\textsuperscript{\rm 1,\rm 4}, Shuya Wang\textsuperscript{\rm 1,\rm 5}, 
}

\centerline{
Xiaofeng Wang\textsuperscript{\rm 1,\rm 6}, Xianghui Ze\textsuperscript{\rm 7}, Yifan Chang\textsuperscript{\rm 2}, Guosheng Zhao\textsuperscript{\rm 1}, Jiangnan Shao\textsuperscript{\rm 1}, Guan Huang\textsuperscript{\rm 1}, Hengyu Liu\textsuperscript{\rm 8},
}

\centerline{
 Yonggang Zhang\textsuperscript{\rm 3}, Wei Xue\textsuperscript{\rm 3}, 
 Chunyuan Guan\textsuperscript{\rm 9}, Chenglin Pu\textsuperscript{\rm 9},
 Yike Guo\textsuperscript{\rm 3}, Xingang Wang\textsuperscript{\rm 2}, Zheng Zhu\textsuperscript{\rm 1 \Letter}
}

\vspace{1em}

\centerline{\textsuperscript{\rm 1}GigaAI~~~~\textsuperscript{\rm 2}University of Chinese Academy of Sciences~~~~\textsuperscript{\rm 3}Hong Kong University of Science and Technology} 
\centerline{~~~~\textsuperscript{\rm 4}University of Leeds~~~~
\textsuperscript{\rm 5}Cornell University~~~~\textsuperscript{\rm 6}Tsinghua University~~~~}
\centerline{\textsuperscript{\rm 7}Nanjing University of Science and Technology~~~~\textsuperscript{\rm 8}The Chinese University of Hong Kong~~~~\textsuperscript{\rm 9}FAWTD}

\vspace{1em}

\centerline{{Project Page: \href{https://open-gigaai.github.io/Zero2Skill}{https://open-gigaai.github.io/Zero2Skill}}} 

\vspace{-1em}
}

\begin{document}
\maketitle

\renewcommand{\thefootnote}{\fnsymbol{footnote}}

\begingroup
\renewcommand{\thefootnote}{*}
\footnotetext{Equal Contribution.}

\renewcommand{\thefootnote}{\Letter}
\footnotetext{Corresponding authors. zhengzhu@ieee.org}
\endgroup

\begin{abstract}
Autonomous data collection governs both the volume and quality of real-world trajectories available for manipulation policy learning.
Existing pipelines reduce human effort through self-resetting mechanisms, VLM-based verification, or language-guided correction.
However, in long-horizon collection, the same failure modes recur across episodes. A correction scoped to the episode at hand must be re-issued at every recurrence, so the cost of oversight grows with session duration rather than with the number of distinct problems.
We present \textsc{\method}, a human-robot symbiotic agentic system in which the robot improves under natural-language guidance. 
The operator supplies corrective knowledge in natural language, and every correction is retained for reuse in all subsequent rounds. 
As this knowledge accumulates, a failure mode that has been corrected once is typically handled automatically thereafter, so the operator's attention is drawn to new problems rather than to recurrences of old ones. 
\textsc{\method} formulates the collection loop as a verification-gated decision process, in which the system collects, verifies, and resets autonomously and pauses to notify a remote operator only when a phase fails verification repeatedly, a boundary set by an explicit retry budget. 
The operator describes the observed problem in natural language.
An LLM parser translates each utterance into a structured system adjustment and stores it in a Corrective Memory consulted on subsequent rounds, so that an addressed failure mode is less likely to require a second correction under the same conditions.
On a real-robot desktop-clearing testbed, \textsc{\method} matches the episode collection success rate of full human teleoperation while reducing human working time to $16\%$ of that required by teleoperation.
Language corrections repair both verifier criteria and execution strategies, improving the verifier's agreement with human labels in all four evaluated settings and raising average single-attempt collection success from $12.5\%$ to $47.5\%$, with a separate arm-selection correction improving it from $20.0\%$ to $50.0\%$.
Closing the loop to deployment, policies fine-tuned on \textsc{\method}-collected data match the policy success rate of those trained on full teleoperation data while requiring only a fraction of the human working time during collection.

\end{abstract}

\abscontent
\section{Introduction}
Vision-Language-Action (VLA) models have emerged as a promising paradigm for general-purpose robotic manipulation, driven by their ability to map raw perception and language instructions directly to motor commands~\citep{rt2_2023, openvla_2024, pi0_2024}.
The efficacy of these models is tightly coupled with the volume and quality of real-world demonstration data available for training~\citep{openx_2024, droid_2024}.
As model capacity continues to grow, data acquisition in the physical world, rather than architectural innovation, has become the primary bottleneck for further progress~\citep{scalingup_2024}.

A broad spectrum of methods has been proposed to address this bottleneck. At one end, teleoperation systems such as ALOHA~\citep{aloha_2023} and UMI~\citep{umi_2024} produce high-fidelity trajectories but require sustained operator attention for every episode. 
At the other end, fully autonomous pipelines employ self-resetting mechanisms~\citep{roboclaw_2026}, VLM-based verification~\citep{radar_2026}, or heuristic scripts to remove human labor entirely, though long-tail physical failures, such as perception drift, object displacement, and strategy deadlocks, cause these systems to degrade silently over extended sessions. Between these extremes, interactive approaches allocate human effort more judiciously. ThriftyDAgger~\citep{thriftydagger_2021} and Fleet-DAgger~\citep{fleetdagger_2022} route operator attention to uncertain states, while language-guided correction methods~\citep{yay_2024, droc_2023, irosa_2026} allow operators to refine robot behavior through natural-language feedback during single-task execution. 

However, long-horizon collection places a demand on human feedback that single-task execution does not. Over a multi-hour session, the same failure mode (a mismatched success criterion, a recurring perception error, an ill-suited grasp strategy) can recur across many episodes, so the value of a correction depends not only on whether it resolves the immediate failure but on whether it remains available the next time the failure appears. Methods designed for execution-time correction naturally scope feedback to the task at hand: interventions in interactive imitation learning are consumed as demonstrations for the next policy update~\citep{thriftydagger_2021,fleetdagger_2022}, and language-correction methods apply feedback to the current episode or retain it for task plans and skill parameters~\citep{yay_2024, droc_2023, irosa_2026}. The collection loop itself—its success criteria, reset checks, and sampling priorities—has, to our knowledge, no comparable retention mechanism. Without one, the cost of oversight accrues with the number of episodes rather than with the number of distinct problems a session presents. If each intervention instead produced a reusable rule that the collector consults in subsequent rounds, the same failure mode would seldom require a second correction: intervention frequency could then decrease over a session, and the system could grow more autonomous through accumulated operational knowledge rather than through model updates. This exchange, in which the operator supplies corrective knowledge in natural language and the system returns steadily growing autonomy, constitutes a human-robot symbiosis. 
Furthermore, modern LLM-based agentic systems~\citep{openclaw,lu2026towards,yang2026clawnet} can interpret open-ended language, orchestrate tools~\citep{ma2026skillclaw}, and maintain persistent memory across sessions~\citep{xu2026mem,zhang2025survey}, providing precisely the substrate needed to retain corrective knowledge, match it against new situations, and apply it without retraining.

\begin{figure}[t!]
    \centering
    \includegraphics[width=0.95\textwidth]{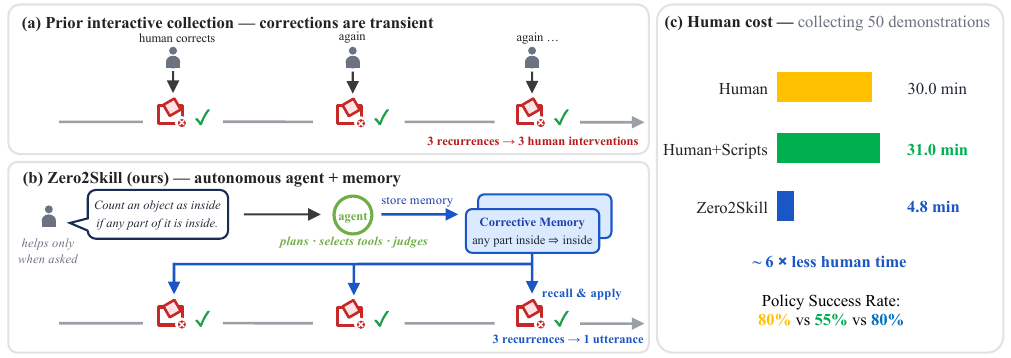}
    \caption{(a) When each correction serves only the current episode, a recurring failure mode requires a new human intervention at every recurrence. (b) \textsc{\method} parses each operator utterance into a structured rule stored in Corrective Memory and reapplied on subsequent rounds, so a single utterance covers later recurrences of the same failure mode. (c) Human cost of collecting 50 valid demonstrations and the resulting policy success rates. }
    \label{fig:teaser}
\end{figure}

Guided by this principle, we present \textsc{\textbf{\method}}, a human-robot symbiotic agentic system that realizes this exchange at two timescales with a strict division of labor.
Within a session, the system adapts its behavior without touching the model.
It collects, verifies, and resets autonomously, each phase gated by an explicit VLM verdict, and pauses to asynchronously notify a remote operator once a phase exhausts its retry budget. 
The operator describes the observed problem in natural language, and an LLM parser translates each utterance into a structured system adjustment, such as sharpening the VLM success verifier's criteria or rebalancing object sampling priorities, persisting it as text in a Corrective Memory consulted on all subsequent rounds.
This mechanism is entirely model-agnostic, modifying neither the underlying VLA weights nor the perception backbone, so its corrections remain valid whenever the underlying policy changes.
Between sessions, the model itself changes.
Trajectories certified under the corrected criteria are consolidated to fine-tune the underlying policy, so human language reaches the weights only indirectly, through the data it certifies; a stronger policy could then re-enter collection with fewer failures and less need for attention, forming a multi-round data flywheel of which this report evaluates a single collect--train--deploy cycle.
In short, within a session the system changes its behavior through text alone, and between sessions it changes its weights through data alone.
We view this architecture as a first step toward a fuller symbiosis in which natural-language feedback serves directly as a learning signal for continual policy updates, with the present system realizing that vision at the data level.
We validate \textsc{\method} on a real-robot pick-and-place testbed spanning multiple object categories. As previewed in Figure~\ref{fig:teaser}, collecting $50$ valid demonstrations requires $4.8$ minutes of human working time with \textsc{\method} versus $30.0$ minutes under full teleoperation, a roughly six-fold reduction, while the policy fine-tuned on \textsc{\method} data matches the $80\%$ policy success rate of the policy trained on teleoperation data. Our main contributions are as follows: 
\begin{enumerate}
    \item We propose \textbf{\textsc{\method}}, a human-robot symbiotic agentic system for embodied data collection that formulates the collection loop as a verification-gated decision process with an explicit intervention boundary. The system operates autonomously over extended sessions and invokes human assistance only when its estimated reliability falls below a tunable threshold, achieving a human-time ratio of 16\% while matching the episode collection success rate of full human teleoperation. 
    
    \item We introduce a general-purpose language feedback channel backed by a persistent \textbf{Corrective Memory}, through which an LLM parser translates operator utterances into structured rules and addresses failures at both the perception and the action level via a single interface.
    Language revisions to the judging criteria improve verifier accuracy in all four evaluated settings, three of which reach $10/10$ while the hardest improves from $0/10$ to $4/10$, language-guided execution corrections raise the average single-attempt collection success rate from $12.5\%$ to $47.5\%$ through cumulative segmentation and grasp-depth corrections, and from $20.0\%$ to $50.0\%$ in a controlled arm-selection comparison.
    
    \item We close the loop from collection to deployment.
    Policies fine-tuned on \textsc{\method}-collected data reach a policy success rate comparable to those trained on full-teleoperation data while requiring substantially less runtime human effort during collection.
    We further introduce \textbf{Trajectories-per-Human-Minute (TpHM)} as an evaluation axis that jointly measures collection efficiency and the utility of the data.

\end{enumerate}

\section{Related Work}

\subsection{Scaling Robot Data Collection}

The demand for large-scale demonstration data has motivated a broad spectrum of collection strategies. At the fully manual end, teleoperation systems such as ALOHA~\citep{aloha_2023} and UMI~\citep{umi_2024} produce high-fidelity trajectories, and cross-institutional efforts including DROID~\citep{droid_2024} and Open X-Embodiment~\citep{openx_2024} have assembled datasets of unprecedented scale. These approaches share a common bottleneck: every trajectory requires sustained operator attention, making collection cost proportional to dataset size.

To reduce this dependence on human labor, recent work explores fully autonomous collection. RoboClaw~\citep{roboclaw_2026} introduces Entangled Action Pairs that couple forward manipulation with inverse recovery to form self-resetting loops, enabling continuous on-policy data acquisition under a VLM-based meta-controller. RADAR~\citep{radar_2026} achieves a similar goal through a four-module pipeline of VLM-guided planning, GNN-based execution, VQA-based verification, and FSM-driven environment reset. While these systems substantially reduce human effort, \citet{scalingup_2024} provide a cautionary result: autonomous imitation learning still faces many of the same environment-design challenges as reinforcement learning, and simply collecting more autonomous data often provides less improvement than collecting additional human demonstrations. The core difficulty is that long-tail physical failures accumulate silently during unattended sessions, and fully autonomous systems lack the ability to diagnose and recover from novel failure modes.

Between these two extremes, interactive approaches allocate human effort selectively. ThriftyDAgger~\citep{thriftydagger_2021} learns a switching policy that solicits operator intervention only at states estimated to be sufficiently novel or risky. Fleet-DAgger~\citep{fleetdagger_2022} extends this idea to multi-robot fleets and proposes Return on Human Effort (ROHE) as a metric for measuring fleet-level efficiency of human supervision. Sirius-Fleet~\citep{siriusfleet_2024} further integrates visual world models for anomaly prediction, automatically adapting its intervention threshold as policy performance improves.
Across these lines of work, the knowledge produced by each human intervention is consumed within the current episode and not retained as an explicit rule, so a structurally similar failure may prompt the operator again. \textsc{\method} is designed to retain such corrections as human-readable text in a lightweight Corrective Memory and to replace continuous teleoperation with sparse natural-language feedback.

\subsection{Language- and Agent-Guided Robot Improvement}

One line of work uses natural language to improve robot behavior at execution time. YAY Robot~\citep{yay_2024} enables real-time adaptation to verbal corrections by fine-tuning a hierarchical policy on logged (observation, correction) pairs, showing that even fine-grained spatial feedback such as ``move a bit to the left'' can be incorporated effectively. DROC~\citep{droc_2023} distills online language corrections into a reusable knowledge base and retrieves relevant past experiences via textual and visual similarity. IROSA~\citep{irosa_2026} adopts a tool-based architecture in which an LLM selects and parameterizes predefined skill-adaptation functions, such as speed adjustment or trajectory correction, maintaining a protective abstraction layer between the language model and robot hardware. These methods demonstrate that language is an effective low-cost supervision channel, but their shared scope is single-task execution rather than long-horizon data collection.

A second and more recent line replaces human language with an agent that improves itself. ENPIRE~\citep{xiao2026enpire} has a coding agent construct automated verification and reset from a one-time human setup, after which it refines its own policy from real-world execution without further human involvement. ASPIRE~\citep{lu2026aspire} has a coding agent debug its own failures and distill the fixes into a growing skill library of reusable code. ABot-Claw~\citep{huo2026abotclaw}, built on the same OpenClaw runtime as our system, targets heterogeneous multi-robot coordination, pairing a visual-centric spatial memory for grounded object and place retrieval with a critic-based feedback loop for online correction and replanning. These systems share our goal of reducing sustained human effort, and some accumulate reusable knowledge as \textsc{\method} does. 
In these systems, interventions are recorded as demonstrations that improve the policy at the next training update: an implicit, weight-level form of retention. Within a session, however, they are not available as explicit rules the collector can consult, so a structurally similar failure may prompt the operator again before the next update. \textsc{\method} complements this line of work by retaining corrections as human-readable text in a lightweight Corrective Memory and by replacing continuous teleoperation with sparse natural-language feedback.

\section{Method}
\label{sec:method}
\begin{figure}[t]
\centering
\includegraphics[width=0.95\linewidth]{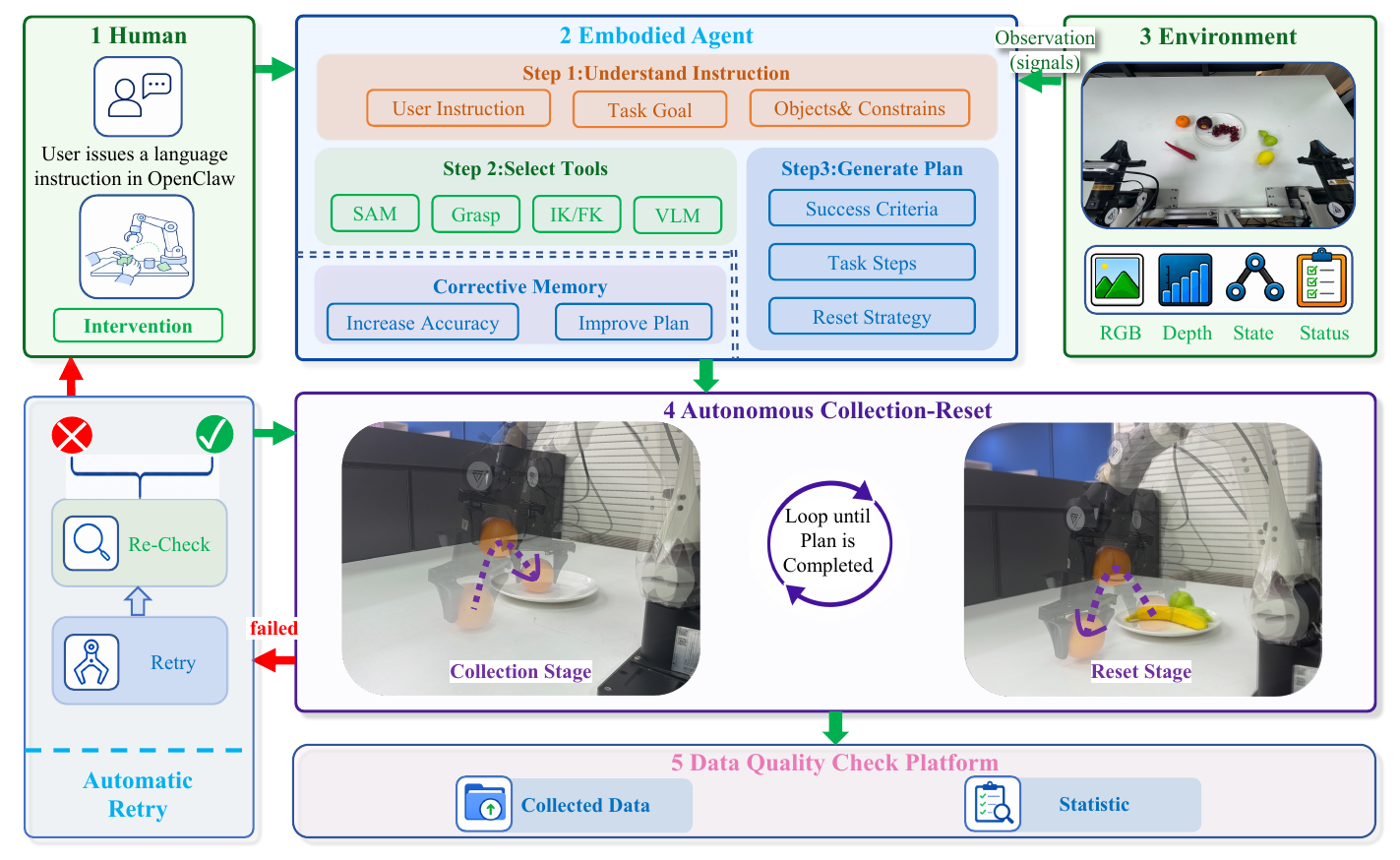}
\caption{Overview of the agent-guided robotic data collection pipeline in \textsc{\method}.}
\label{fig:pipeline}
\end{figure}

\textsc{\method} is built on OpenClaw~\citep{openclaw} and reuses its skill, tool-calling, and persistent-memory interfaces. As shown in Fig.~\ref{fig:pipeline}, a run begins with a natural-language instruction from the operator. The agent combines the instruction with observations of the environment, analyzes the task, selects feasible tools or APIs, and generates a self-collection plan consisting of task information, a collection plan, and a reset plan. The operator reviews and confirms the plan before execution.

After confirmation, \textsc{\method} enters an autonomous collection--reset loop. In each iteration, the robot performs the collection action, evaluates the result using observation signals such as RGB images, depth, robot state, and task status, and then resets the environment for the next round. Recoverable failures are retried automatically, while repeated failures or disturbed scenes trigger human intervention. A \emph{language intervention} provides remote guidance or correction, whereas a \emph{physical intervention} manually recovers the environment. \emph{Experience memory} comprises Corrective Memory (Section~\ref{subsec:intervention}) together with per-episode logs: the former stores reusable rules and the latter records every successful round and every failure-recovery event for audit and replay. Plan confirmation, language intervention, physical intervention, and monitoring occur at different stages and incur different human costs; all of them are counted toward the human working time reported in Section~\ref{sec:exp}. 

\subsection{Autonomous Collection}
\label{subsec:collection}

\noindent\textbf{Task analysis and planning. }
Given an instruction such as \textit{``learn to tidy the tabletop''}, the agent produces a complete plan in a single pass. It queries the camera for an observation, extracts the scene and the objects present, and uses the task description together with the tool library to determine the manipulation goal and the tool-call arguments for collection and reset. We describe the concrete manipulation tools in Appendix~\ref{app:tools}. The resulting plan comprises four parts: a collection routine that specifies the per-episode action chain and its tool calls, a collection success criterion expressed as a verification prompt, a reset routine that returns the scene to its initial state, and a reset success criterion. These criteria are generated by the agent rather than hand-specified, and are therefore fallible; the language intervention mechanism of Section~\ref{subsec:intervention} corrects them, and the persistent corrections described there are edits to the prompts the agent authors here.

\noindent\textbf{The collect-and-reset loop. }
Collection then runs as a closed control loop, given in Algorithm~\ref{alg:loop}. To gather a target of $K$ trajectories, the loop alternates a collection phase and a reset phase, each driven by its own command ($c_{\mathrm{col}}, c_{\mathrm{res}}$) and checked by its own verification prompt ($p_{\mathrm{col}}, p_{\mathrm{res}}$). A phase executes its command, waits for the scene to settle, and queries a VLM on a fresh image~\citep{nils_2024}. The prompt states the success criterion of the phase and instructs the model to begin its reply with an explicit verdict, \textsc{yes} or \textsc{no}, followed by a one-sentence justification. The phase passes on a \textsc{yes} verdict and is retried otherwise, up to a per-phase limit $N$ (set to three by default). A reply that carries no parseable verdict, including one produced by an infrastructure error such as a timed-out model call, never passes the check; Section~\ref{subsec:labeling} describes how such replies are kept out of the trajectory labels. When a phase exhausts its retries, the loop halts the arm, enters the \textsc{Alert} state, and notifies the operator. Rather than crashing or retrying blindly, it waits for the operator's feedback and then resumes the phase, so that collection continues from a corrected state rather than a degrading one.
The retry limit $N$, together with the strictness of the verification criteria, determines where the system places the boundary between operating autonomously and requesting help. A scripted collector leaves this boundary implicit~\citep{roboclaw_2026, radar_2026}, fixed by whatever the control logic does upon a failed check. \textsc{\method} instead makes both parts of the boundary explicit. The retry limit is a hyperparameter set prior to a run: a lower $N$ makes the system more inclined to stop and ask, preserving data quality at the cost of more frequent interruptions, while a higher $N$ has the opposite effect. The criteria are plain text authored at planning time, so their strictness can also be recalibrated during a run through the language channel of Section~\ref{subsec:intervention}: a criterion that rejects attempts a human would accept drains the retry budget on false alarms, and a single corrective sentence stops that loss for every subsequent round. Making the boundary explicit allows this trade-off to be set deliberately rather than inherited from an implementation.

\begin{algorithm}[t]
\caption{Collect-and-Reset Loop}
\label{alg:loop}
\begin{algorithmic}[1]
\REQUIRE collection command $c_{\mathrm{col}}$, reset command $c_{\mathrm{res}}$, verification prompts $p_{\mathrm{col}}, p_{\mathrm{res}}$, target count $K$, retry limit $N$
\STATE $k \gets 0$
\WHILE{$k < K$}
    \STATE \textsc{RunPhase}($c_{\mathrm{col}}, p_{\mathrm{col}}$) \COMMENT{collection phase}
    \STATE record trajectory; \quad $k \gets k + 1$
    \STATE \textsc{RunPhase}($c_{\mathrm{res}}, p_{\mathrm{res}}$) \COMMENT{reset phase}
\ENDWHILE
\vspace{3pt}
\STATE \textbf{procedure} \textsc{RunPhase}($c, p$)
\STATE \quad \textbf{loop}
\STATE \quad\quad \textit{ok} $\gets \textsc{false}$
\STATE \quad\quad \textbf{for} $\textit{attempt} = 1$ \textbf{to} $N$ \textbf{do}
\STATE \quad\quad\quad execute $c$; wait for the scene to settle
\STATE \quad\quad\quad \textbf{if} $\textsc{Verify}(p) = \textsc{yes}$ \textbf{then} \textit{ok} $\gets \textsc{true}$; \textbf{break} \COMMENT{verdict $\in$ \{\textsc{yes}, \textsc{no}, \textsc{unknown}\}}
\STATE \quad\quad \textbf{if} \textit{ok} \textbf{then return}
\STATE \quad\quad enter \textsc{Alert}; apply operator feedback \COMMENT{then retry the phase}
\end{algorithmic}
\end{algorithm}

\noindent\textbf{Why the reset is verified separately. }
The reset check warrants particular attention. A failed grasp is self-evident, leaving the gripper empty so that the round simply repeats, whereas a failed reset is silent: it leaves the scene slightly altered, and in the absence of a check the system would continue recording from a configuration that drifts progressively further from its intended state until no grasp can succeed. Verifying the reset against its own criterion is what allows the system to detect this drift and pause before a long run degrades unnoticed.

\subsection{Language Intervention and Corrective Memory}
\label{subsec:intervention}
In the \textsc{Alert} state the operator sends a free-form sentence over the OpenClaw channel. This is the language intervention of Section~\ref{sec:method}, with physical recovery reserved for the rare scene that cannot be reset in software. An LLM parser interprets the sentence together with the current state and the current memory, and decides whether the correction should persist.

The parser distinguishes two kinds of correction. Some apply to every subsequent instance of the same problem. Telling the system that a small offset still counts as success is one such case, which the parser appends to the relevant verification prompt. Telling it that soft, deformable objects require a firmer grasp than rigid ones is another, which the parser encodes as a conditioned rule. Both are written to Corrective Memory and consulted on every subsequent round, after which the failure they address no longer raises an alert. Other corrections apply only to the current round. An instruction such as ``nudge it slightly to the left this time'' frees a stuck grasp but carries no implication for the next object, so the parser applies it once and stores nothing. Corrections of the first kind are intended to reduce intervention frequency over a session, while those of the second only restart a stalled run. The conditioning is what makes the first kind reusable: an instruction such as ``go down one centimeter'' cannot accumulate on its own, because nothing records the condition under which it should apply, but once conditioned, as in ``for flat, low-profile objects, grasp one centimeter below the height proposed by the policy,'' it generalizes to every object that matches. The system does not, on its own, generalize from a single instruction to a rule covering other objects; the scope of a correction is always either stated by the operator or inferred by the LLM for the object at hand.

\noindent\textbf{A general-purpose channel. }
The channel presupposes nothing about the content of operator feedback. An incorrect success criterion, an object that should be skipped, a grasp that requires a condition, and a scene that requires a reset are all handled through the same parser. The range of adjustments available to the operator is set by the parser's capacity to interpret language, rather than by a predefined list of exposed parameters. 

\noindent\textbf{Corrective Memory. }
Corrective Memory is a lightweight store of persistent corrections, kept as a structured Markdown file that the operator can read and edit. Each entry has four fields: a \emph{trigger} that specifies when it applies, a \emph{correction} that specifies the adjustment, a \emph{scope} that bounds the trigger, and the \emph{source utterance}, kept verbatim for audit. The deformable-object remark above, for example, becomes an entry whose trigger is a soft or deformable object, whose correction increases the grasp force, and whose scope is the grasp phase. At the start of each round the system consults the store and applies any entry whose trigger matches the current situation. Entries are grouped by the component they act on, namely verification prompts, conditioned grasp policies, object priorities, and retry limits, and further groups can be added without changing this procedure. 
% The parser interface, the storage format, and the consultation points of Corrective Memory are specified in Appendix~\ref{app:memory}.

\subsection{Trajectory Labeling and Quality Certification}
\label{subsec:labeling}
Verification decides whether the loop proceeds. It does not by itself produce a dataset. A labeling layer sits between the collection loop and policy learning. Each episode is saved in its own timestamped session folder with its logs and a rendered multi-view video. Each episode receives two labels. An \emph{execution label} records whether the action chain ran to completion. An \emph{outcome label} records whether the task succeeded, judged by the VLM verifier as success, failure, or unknown. Tracking the two separately shows whether a problem came from the tooling or from the policy.

Errors from the infrastructure are kept out of the labels. If the verifier output shows signs of an infrastructure error, such as a dropped connection or a timed-out model call, the episode is marked unknown instead of failed. A preflight check also requires every camera and joint topic to publish a message before the arm moves, so a hung sensor cannot produce an empty episode. Since each episode keeps its final observations, it can be re-judged offline whenever a verification prompt is corrected (Section~\ref{subsec:intervention}), so the whole dataset stays labeled under one standard.

After each episode, the system computes quality statistics, such as motion smoothness and image sharpness, and collects them into a report the operator can read remotely. A quality check then rates each recording as good, marginal, or bad. This catches problems like an arm that never moved, a dark camera, or jerky motion. The rating is independent of task success. A failed grasp can still be a well-recorded and useful episode.
At training time, an episode is kept only if it ran to completion, passed the outcome rule, and is not flagged as a bad recording. Kept episodes are merged into the training format and listed in a fixed manifest, so every training set in Section~\ref{sec:exp} can be reproduced exactly. The rest of the pipeline is automatic. Training runs on a remote server while collection continues, and each checkpoint is evaluated on the robot under the protocol of Appendix~\ref{app:train-eval}, with unknown counted as failure.

\subsection{Policy Learning and the Data Flywheel}
\label{subsec:train}
The trajectories that pass verification are consolidated into a dataset and used to fine-tune the underlying VLA policy~\citep{pi0_2024,openvla_2024}. One aspect of this consolidation is consequential for the method: each trajectory is retained or discarded according to the collection success criterion \emph{in the state the operator left it} after the corrections made during collection. A criterion that began too strict or too lenient has by this point been adjusted through language, so that the quality of the training set inherits every persistent correction made during the run. This is the link between the language channel and the data, in that the same utterances that can reduce intervention frequency also govern which trajectories are deemed clean. The resulting policy is then evaluated on the physical platform in a deployment setting, with no agent and no verifier in the loop, so that its policy success rate reflects the collected data rather than the surrounding support. We defer the fine-tuning and evaluation protocols, which follow standard practice, to Appendix~\ref{app:train-eval}.

Collection, language feedback, and policy improvement are designed to form a self-reinforcing loop across multiple runs: a policy trained on one session can serve as the collector for the next, a stronger collector can succeed more often unaided, and persistent corrections in Corrective Memory can continue to apply so that human involvement diminishes as the loop proceeds.
In the experiments reported here we evaluate one complete cycle---collect, correct, consolidate, fine-tune, and deploy---and leave full multi-round flywheel validation to future work.

%===============================================================================

\section{Experiments}
\label{sec:exp}

We design experiments to answer three questions: (1)~Does interactive autonomy reduce the need for continuous human teleoperation? (2)~Does language-guided correction recover the system from long-tail failures? (3) Is the data collected by \textsc{\method} good enough, i.e., does it train a deployed policy on par with one trained on human teleoperation data?

\subsection{Experimental Setup}
\label{subsec:exp-setup}
\noindent\textbf{Platform and task.}
All experiments are conducted on a dual-arm Piper setup with a head-mounted RGB-D camera. We study a single \emph{desktop-clearing} task, in which the robot clears target objects from the tabletop into a container.

\noindent\textbf{Policy fine-tuning.} We fine-tune from $\pi_{0.5}$~\citep{pi05} following the training and evaluation protocol in Appendix~\ref{app:train-eval}.

\noindent\textbf{Evaluation metrics.}
We report four outcome metrics throughout this section, each answering a different question about system performance.
\emph{Episode collection success rate} is the fraction of attempted episodes for which the collect--verify--reset loop ultimately produces a valid trajectory, counting automatic retries and sparse human intervention within the episode (Experiment~1).
\emph{Single-attempt collection success rate} is the fraction of individual collection tries that succeed on the first execution, without counting within-episode retries or intervention; language-guided corrections are evaluated on this metric in Experiment~2.
\emph{Verifier accuracy} is the fraction of collection or reset judgments on which the VLM verifier agrees with human-provided success/failure labels (Experiment~2).
\emph{Policy success rate} is the fraction of deployment trials in which a fine-tuned policy completes the task with no agent, verifier, or human support at test time (Experiment~3).
We also report \textbf{Trajectories-per-Human-Minute (TpHM)}, defined as
$\mathrm{TpHM} = (\text{number of successful episodes}) / (\text{human working time in minutes})$,
which jointly measures collection throughput and the human effort required to obtain it. 

\noindent\textbf{Models and baselines.}
The VLM verifier is Seed1.8~\citep{seed2026seed18modelcardgeneralized} and the LLM parser is DeepSeek-V3.2~\citep{deepseekai2025deepseekv32pushingfrontieropen}. The scripted baseline is a shell script that loops predetermined collection and reset routines without verification, automatic retries, or language intervention. An operator remains on site throughout its runs to monitor execution and recover from failures manually.

\noindent\textbf{Verifier and confidence gating.} 
At each phase the verifier is queried on a fresh image and instructed to begin its reply with an explicit verdict, \textsc{yes} or \textsc{no}, followed by a one-sentence justification. A phase passes on a \textsc{yes} verdict and is otherwise retried up to $N=3$ times, after which the loop enters the \textsc{Alert} state and requests human input. A reply without a parseable verdict, including one caused by an infrastructure error, does not pass the check and is recorded as \emph{unknown} in the episode label. 
% Prompt templates and the parsing rules are given in Appendix~\ref{app:verification}.

\begin{table}[t]
\centering
\caption{Collection efficiency for Desktop Clearing (50 attempted episodes per mode). \emph{Human teleoperation time} counts continuous manual control only. \emph{Human working time} counts total operator attention, including teleoperation, interventions, recovery, and monitoring. \emph{Episode collection success rate} is the fraction of attempted episodes that ultimately yield a valid trajectory. TpHM is successful episodes per minute of human working time. }
\label{tab:collection_comp}
\resizebox{0.82\linewidth}{!}{
\begin{tabular}{@{}lccc@{}}
\toprule
\textbf{Metric} & \textbf{Human Teleop} & \textbf{Scripts} & \textbf{\textsc{\method} (Ours)} \\ \midrule
Attempted episodes & 50 & 50 & 50 \\
Successful episodes & 50 & 35 & 50 \\
Failed episodes & 0 & 15 & 0 \\
Total wall-clock time (min) & 35.0 & 31.0 & 53.0 \\
Human teleoperation time (min) & 30.0 & 1.0 & 0.0 \\
Human working time (min) & 30.0 & 31.0 & 4.8 \\
Episode collection success rate (\%) & 100.0 & 70.0 & 100.0 \\
TpHM $\uparrow$ & 1.67 & 1.13 & \textbf{10.42} \\
\bottomrule
\end{tabular}}
\end{table}

\begin{figure}[t]
\centering
\includegraphics[width=0.95\linewidth]{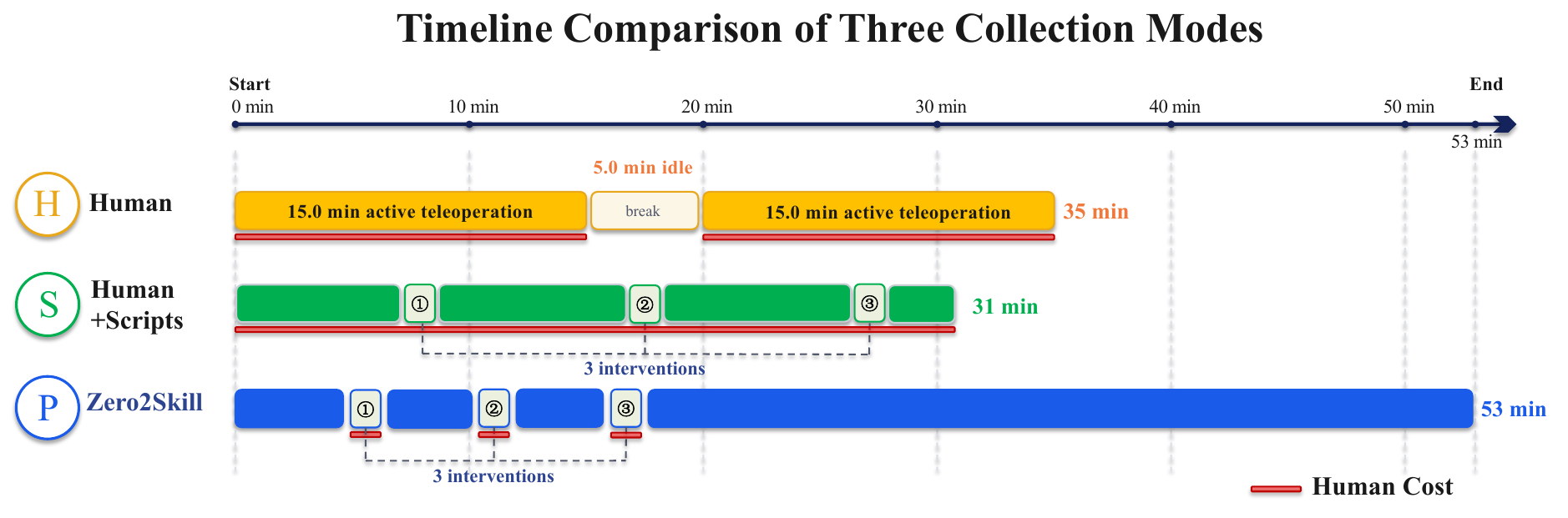}
\caption{Timeline comparison of the three collection modes.}
\label{fig:collection-timeline}
\end{figure}

\subsection{Experiment 1: Collection Efficiency \& Human Cost}
\label{subsec:exp-efficiency}
We first measure the physical effort required to collect demonstrations for a ''Desktop Clearing'' task. The goal of this experiment is not only to measure wall-clock collection time, but also to quantify how much of that time must be actively spent by a human operator. We compare three collection modes: full human teleoperation, a scripted autonomous collector, and \textsc{\method}.

All three modes use the same task definition, workspace, object set, and final success criterion. The human baseline records trajectories by continuous teleoperation. The scripted baseline loops the same collection and reset scripts in a shell without verification or automatic retries; it may finish quickly while still producing failed or invalid episodes, and an operator must monitor the run continuously to handle failures manually. \textsc{\method} runs the autonomous collect--verify--reset loop and only asks for human input when the system requires clarification or recovery. Throughout Section~\ref{sec:exp}, \emph{human teleoperation time} counts only minutes of continuous manual robot control, whereas \emph{human working time} counts the operator's total active attention, including teleoperation, plan confirmation, language interventions, physical recovery, and monitoring during autonomous execution. We report attempted episodes, successful episodes, failed episodes, total wall-clock time, both human-time metrics, episode collection success rate, and TpHM.

Table~\ref{tab:collection_comp} and Figure~\ref{fig:collection-timeline} compare human teleoperation, scripted collection, and \textsc{\method} on the Desktop Clearing task. All three modes attempt the same number of episodes $(50)$. The scripted collector finished in $31.0$ minutes of wall-clock time, of which $35$ episodes were successful and 15 failed, yielding a $70.0\%$ episode collection success rate. Because the scripted run has no verifier or automatic retry, the operator monitored it continuously for the full $31.0$ minutes and handled failures manually, so its human working time equals its wall-clock time. Full human teleoperation successfully completed all $50$ episodes, requiring $30.0$ minutes of continuous teleoperation and $30.0$ minutes of total human working time; the remaining $5.0$ minutes of wall-clock time covers operator breaks between episodes. \textsc{\method} also achieved a $100.0\%$ episode collection success rate across $50$ episodes with $0.0$ minutes of teleoperation and only $4.8$ minutes of human working time. Although \textsc{\method} required a longer total wall-clock time ($53.0$ minutes), it reduced both teleoperation and total human working time while maintaining reliable data collection through verification and selective intervention. \textsc{\method} attains a TpHM of $10.42$, roughly six times that of teleoperation, while matching its episode collection success rate.

\begin{table}[tbp]
\centering
\caption{Language-guided correction cases and VLM verifier accuracy. Each cell reports correct judgments / total cases.}
\label{tab:language_correction}
\resizebox{0.98\linewidth}{!}{
\begin{tabular}{@{}p{0.28\linewidth}p{0.20\linewidth}p{0.12\linewidth}ccc@{}}
\toprule
\textbf{Task Setting} & \textbf{Correction Focus} & \textbf{Judg. Type} & \textbf{Before Correction} & \textbf{After Correction} & \textbf{Accuracy Gain} \\ 
\midrule
Basket desktop clearing & Partial & Collection & 4 / 10 (40.0\%) & 10 / 10 (100.0\%) & +60.0 \\
Basket desktop reset & Relaxed Reset & Reset & 0 / 10 (0.0\%) & 10 / 10 (100.0\%) & +100.0 \\
Box fruit clearing (harder) & Ambiguous & Collection & 0 / 10 (0.0\%) & 4 / 10 (40.0\%) & +40.0 \\
Two-basket object sorting & Category-wise  & Collection & 2 / 10 (20.0\%) & 10 / 10 (100.0\%) & +80.0 \\
\bottomrule
\end{tabular}}
\end{table}

\subsection{Experiment 2: Language-Guided Correction}
\label{subsec:exp-language} 
\begin{figure}[t]
\centering
\includegraphics[width=0.85\linewidth]{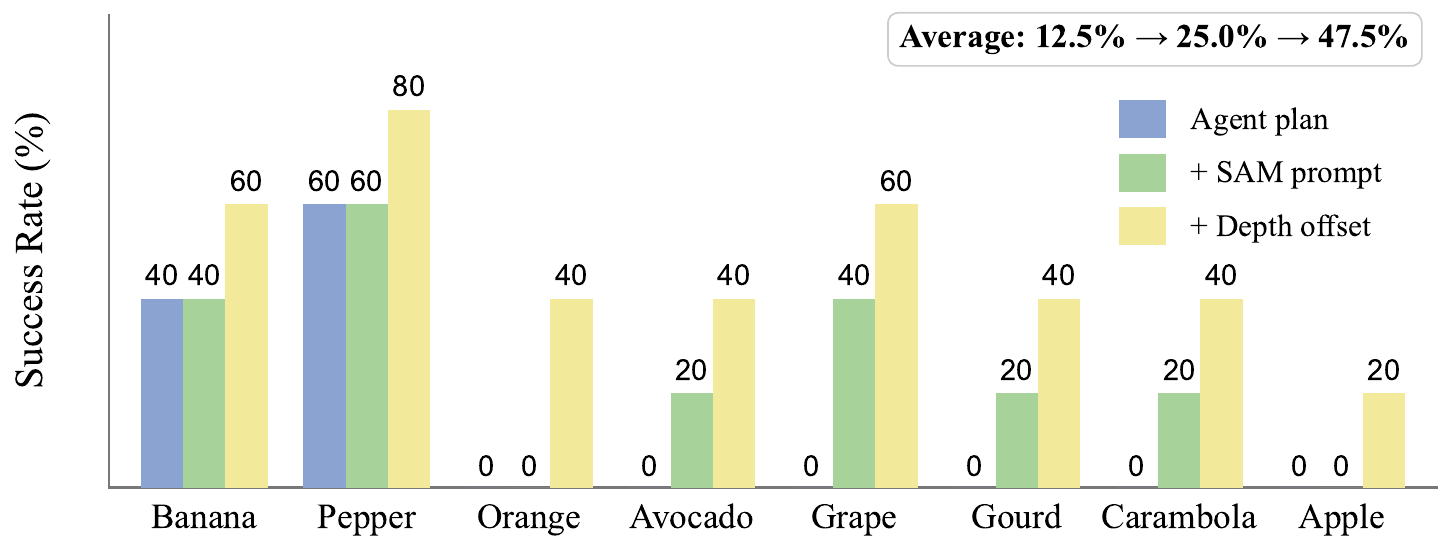}
\caption{Single-attempt collection success rates across eight object categories under cumulative language-guided corrections. Agent plan is the rate under the agent's initial plan, and each subsequent setting adds one correction on top of the previous one: re-prompting the SAM3 segmentation (+SAM prompt) and adjusting the approach depth offset (+Depth offset). Each rate counts only the first collection try per episode, without within-episode retries, and is computed from five attempts per object per setting. }
\label{fig:language-depth-sam}
\end{figure}

\begin{figure}[t]
\centering
\includegraphics[width=0.8\linewidth]{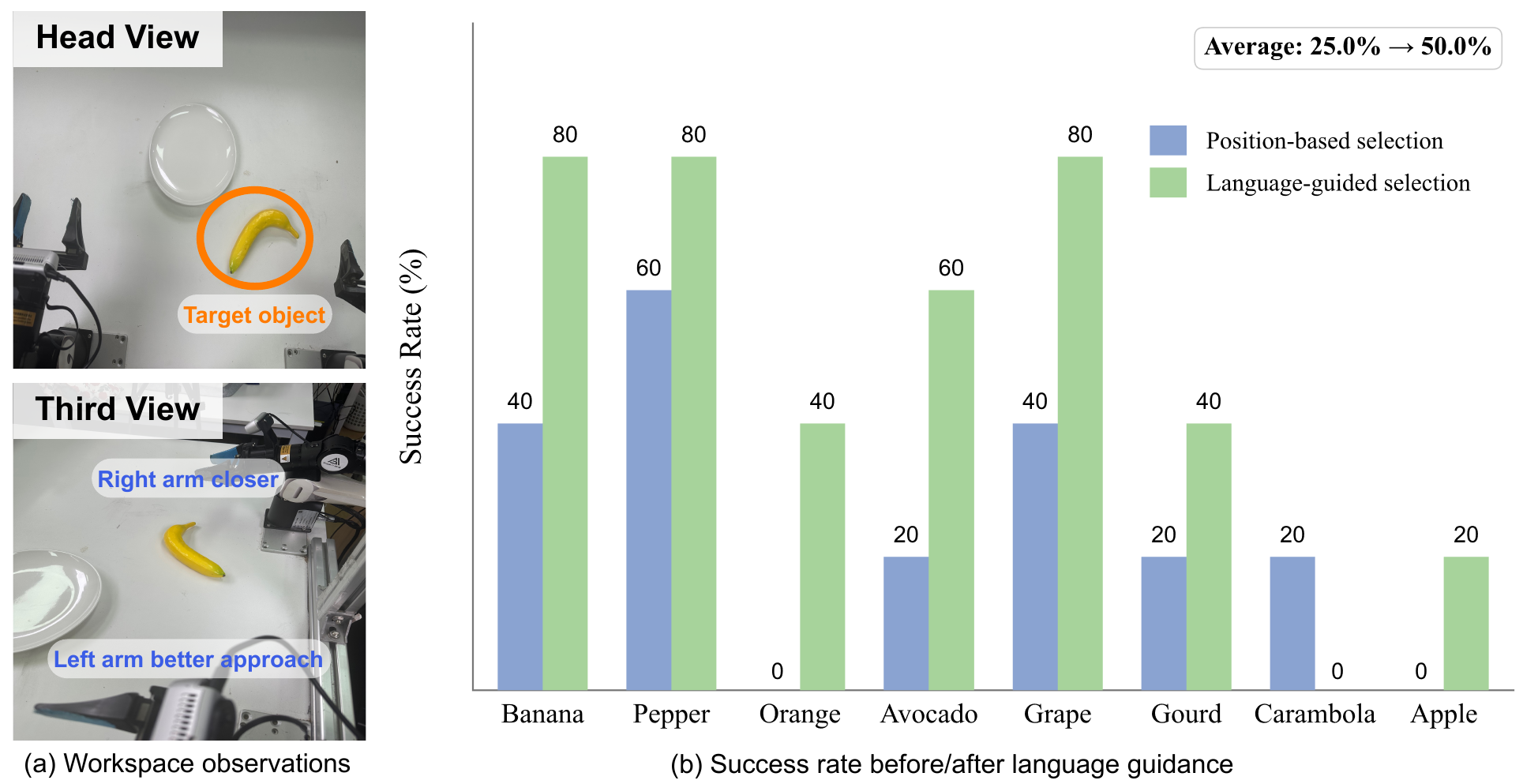}
\caption{Representative observations and single-attempt collection success rates for language-guided arm selection. (a) shows a case where the object is spatially closer to the right arm, but the head and third-person views reveal that the left arm has a clearer and more feasible grasping approach. (b) shows that language guidance corrects this proximity-biased arm choice and improves single-attempt collection success across most objects. }
\label{fig:language-arm-side}
\end{figure}

The second experiment tests whether the language channel serves as an online correction mechanism rather than only as an interface for stopping the robot. The persistent corrections studied here were issued during live collection; those active during the 50-episode collection session of Experiment~1 (Section~\ref{subsec:exp-efficiency}) are therefore inherited by the training data of Experiment~3 (Section~\ref{subsec:exp-quality}). This experiment measures success at a finer granularity than Experiment~1. Under the episode-level metric of Experiment~1, an episode counts as successful once the collect--verify--reset loop eventually produces a valid trajectory, including after automatic retries and sparse language intervention; this is why \textsc{\method} reaches $100.0\%$ in Table~\ref{tab:collection_comp} despite imperfect single tries. The present experiment instead evaluates each correction directly, using \emph{verifier accuracy} for judgment-side corrections and \emph{single-attempt collection success} for execution-side corrections (Section~\ref{subsec:exp-setup}). The analysis proceeds in four parts: a verifier-side evaluation of corrected judging criteria (Table~\ref{tab:language_correction}), two controlled execution-side evaluations that isolate the segmentation and grasp-depth corrections (Figure~\ref{fig:language-depth-sam}) and the arm-selection correction (Figure~\ref{fig:language-arm-side}), and a whole-process analysis of interventions accumulating during live collection (Table~\ref{tab:success-rates-objects-interventions}).

\noindent\textbf{Verifier-side correction.}
The VLM verifier can fail when the automatically generated prompt does not match the human-defined success criterion. In the desktop-clearing task, the initial prompt required all visible objects to be fully placed inside the basket and required the reset stage to restore the original object positions. The human criterion instead counted collection as successful when every object was at least partially inside the container, and reset as successful when at least one object was outside it. This mismatch caused repeated verifier errors. To test whether language feedback corrects it, we compare two settings on the same collection and reset cases. In the no-language setting, the verifier uses the initial prompt. In the language-correction setting, the operator revises the judging rule by specifying that an object counts as inside the container if any part of it is inside, and that reset does not require the exact original positions; the revised prompt is then used for all subsequent judgments. We further evaluate the corrected judging rules on two additional layouts, blue-box clearing and two-basket object sorting. Representative scenes for the four settings are shown in Figure~\ref{fig:basket-verifier-cases} of Appendix~\ref{app:exp}. 
Table~\ref{tab:language_correction} summarizes the results. On the basket desktop task, accepting partial containment raises collection verifier accuracy from $40.0\%$ to $100.0\%$, and relaxing the reset criterion raises reset verifier accuracy from $0.0\%$ to $100.0\%$. The remaining rows evaluate task-dependent judging rules. Category-wise sorting reaches $100.0\%$ after correction, while the harder blue-box case improves from $0.0\%$ to $40.0\%$, indicating that ambiguous containment cases can require iterative correction.

\noindent \textbf{Execution-side correction: segmentation and grasp depth. }
We next test whether language feedback improves the collector's raw execution, independent of verifier judgments. Two persistent corrections are evaluated cumulatively, each parsed into a Corrective Memory entry and applied on all subsequent rounds: a revised SAM3 segmentation prompt and an approach-depth offset. Starting from the agent's initial plan, we measure single-attempt collection success per object category after each addition, with five attempts per object per setting.
Figure~\ref{fig:language-depth-sam} shows that both corrections improve raw execution. The average single-attempt success rate rises from $12.5\%$ under the agent's plan to $25.0\%$ after re-prompting SAM3~\citep{sam3}, and to $47.5\%$ after additionally adjusting the approach depth offset. In absolute terms, 5, 10, and 19 of 40 first attempts succeed under the three settings. The two corrections act differently. The depth offset yields the larger average gain ($+22.5$ points) and improves every object category, indicating that the agent's initial approach depth was a systematic rather than object-specific error. SAM re-prompting adds $+12.5$ points concentrated on four categories (avocado, grape, gourd, and carambola), consistent with segmentation errors affecting only a subset of objects. 

\begin{table}[H]
\centering
\caption{Collection success rates (\%) across four scene configurations as language interventions accumulate during live collection. Each intervention is stored in Corrective Memory and applied to all subsequent attempts. The entry at $k$ interventions is the cumulative success rate over all attempts from the start of the run until the next intervention is issued (end of run for the last entry).}
\resizebox{0.55\textwidth}{!}{
\begin{tabular}{lccccc}
\toprule
\multirow{2}{*}{\textbf{Object}} & \multirow{2}{*}{\textbf{Exp. ID}} & \multicolumn{4}{c}{\textbf{Number of Interventions}} \\
\cline{3-6}
 & & \textbf{0} & \textbf{1} & \textbf{2} & \textbf{3} \\
\midrule
Banana        & 1 & 57.14 & 61.11 & --    & --    \\
Banana \& Pepper & 2 & 42.86 & 47.50 & 46.00 & --    \\
Banana \& Grape  & 3 & 60.87 & 66.67 & --    & --    \\
Banana \& Chili  & 4 & 80.00 & 70.59 & 75.86 & 76.47 \\
\bottomrule
\end{tabular}
}
\label{tab:success-rates-objects-interventions}
\end{table}
\noindent\textbf{Execution-side correction: arm selection.}
The third persistent correction addresses a proximity bias in the agent's plan. By default the agent assigns the arm closest to the target, although for some placements the head-mounted and third-person views reveal that the farther arm affords a clearer and more feasible grasping approach (Figure~\ref{fig:language-arm-side}a). We evaluate this correction in a separate controlled comparison: both settings apply the revised SAM3 prompt with the depth offset disabled, so the arm-selection rule is the only difference, again with five attempts per object per setting.
Figure~\ref{fig:language-arm-side}b shows that language-guided arm selection raises the average single-attempt success rate from $20.0\%$ to $50.0\%$ ($+30.0$ points), improving seven of the eight categories. Carambola is the exception and drops from $20.0\%$ to $0.0\%$: for this object the position-based default happens to select the better arm, and the persistent rule overrides it. This comparison illustrates both the strength and the limit of a persistent correction: it is applied automatically to every subsequent attempt, but its benefit is conditional on the rule matching the failure cause of the object at hand.

\noindent\textbf{Corrections in the collection loop.}
The controlled evaluations above isolate one correction at a time. We finally examine how corrections behave when they accumulate during live collection. Table~\ref{tab:success-rates-objects-interventions} reports four collection groups run under the standard protocol, with language interventions issued as failures arise. Each intervention is stored in Corrective Memory and applied to all subsequent attempts. The entry at $k$ interventions is the cumulative success rate over all collection attempts from the start of the run until the next intervention is issued, or until the run ends for the last entry. Because entries are cumulative, attempts from earlier windows remain in the denominator, so the change between successive entries understates the marginal effect of the most recent correction. Even so, the final cumulative rate ends above the pre-intervention level in three of the four groups (banana, $+4.0$ points; banana and pepper, $+3.1$; banana and grape, $+5.8$). The banana-and-chili group ends below it: the cumulative rate falls from $80.00\%$ to $70.59\%$ after the first intervention and recovers only partially thereafter ($76.47\%$). A stored correction is therefore guaranteed to be reused, but not guaranteed to help. Its value depends on whether it matches the actual failure cause, and in harder scenes, where failures may also stem from limitations of the grasping tool or the execution stack, that cause is more difficult to identify.

\subsection{Experiment 3: Data Quality via Downstream Policy}
\label{subsec:exp-quality}
The third experiment asks whether the data collected by \textsc{\method} is good enough to train a competitive policy. Because autonomous collection could in principle trade data quality for reduced human effort, we use \emph{policy success rate} as a direct measure of data quality: if a policy trained on \textsc{\method} data reaches a policy success rate on par with one trained on human teleoperation data, the collected data is of comparable quality. This evaluation removes all collection-time support at test time: the fine-tuned policy runs directly on the physical robot with no agent planner, no VLM verifier, no corrective memory lookup, and no human intervention during an episode. A trial is counted as successful only if the policy completes the task from the initial scene within the fixed horizon.
 
\begin{figure}[H]
\centering
\includegraphics[width=\linewidth]{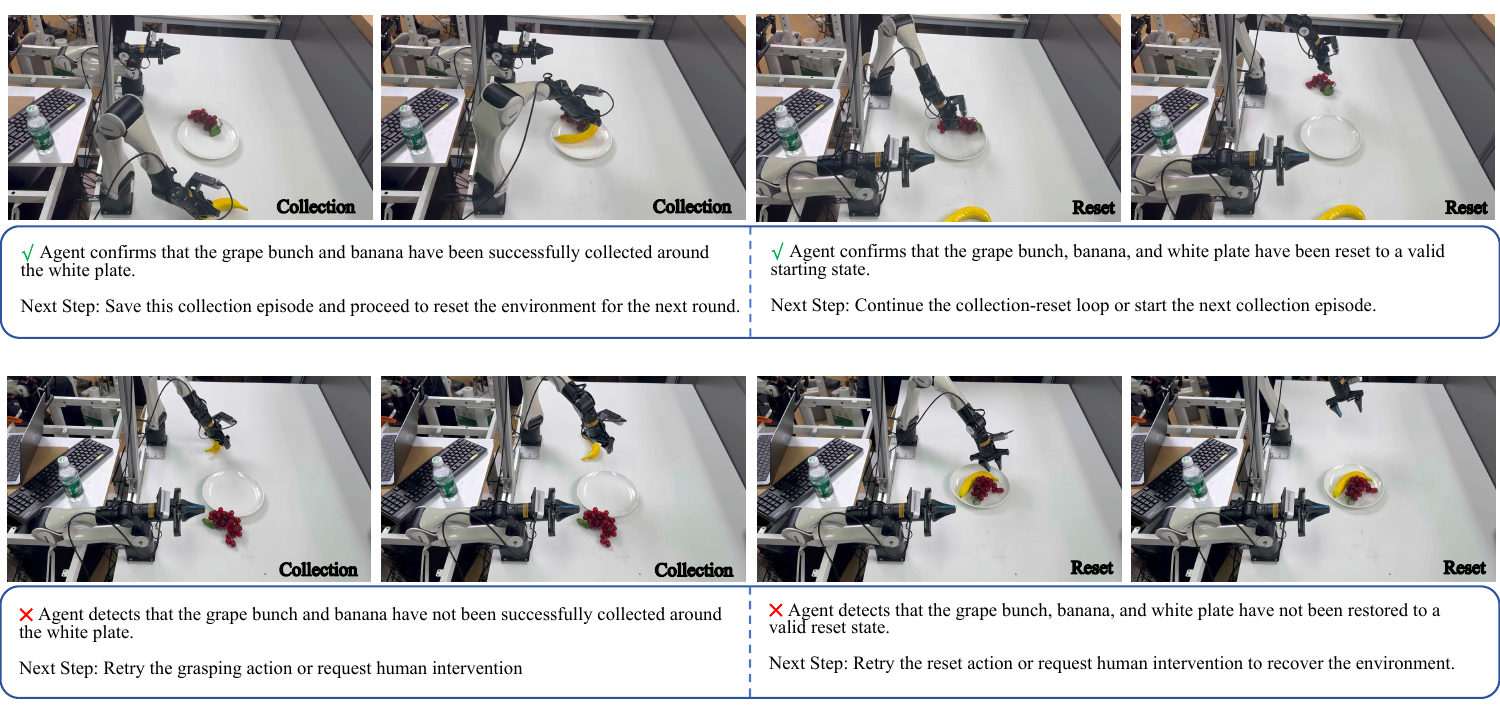}
\caption{Agent feedback for collection and reset verification. The system checks whether the grape bunch, banana, and white plate satisfy the predefined collection or reset criteria. Successful episodes proceed to the next step, while failed episodes trigger a retry or request human intervention.}
\label{fig:collection-reset-feedback}
\end{figure}

Before the resulting trajectories are used for downstream policy training, the agent verifies collection and reset outcomes against predefined criteria. Figure~\ref{fig:collection-reset-feedback} illustrates representative feedback for both successful and failed collection-reset episodes.

To isolate quality from quantity, each policy is fine-tuned on the same number of valid demonstrations ($50$), varying only the collection source. The \textit{Full Human Teleop Data} setting uses $50$ demonstrations recorded by continuous teleoperation. The \textit{\textsc{\method} Data} setting uses the same $50$ verified trajectories from the Experiment~1 collection session; the language-guided corrections analyzed in Experiment~2 were applied during this session, so every persistent rule in Corrective Memory was already in effect when those trajectories were recorded. 
The \textit{Scripted Data} setting uses $50$ valid trajectories from the scripted collector. Because its valid rate is $70\%$, the collector runs additional attempts (about $72$ attempted episodes in total) to reach $50$ valid demonstrations. Since the scripted protocol requires continuous on-site monitoring, this corresponds to roughly 44 minutes of operator time, extrapolated from Table~\ref{tab:collection_comp}. 
All policies are fine-tuned from $\pi_{0.5}$~\citep{pi05} with the same training recipe and are evaluated on the same desktop-clearing task with 20 blind trials each. 

\begin{table}[t]
\centering
\caption{Data-quality diagnostics comparing $50$ full-human-teleoperation episodes with $50$ \textsc{\method}-collected episodes; values are mean $\pm$ sample standard deviation across episodes. Smoothness and Aesthetics are $0--100$ scores, Brightness is grayscale intensity on $0--255$, and RMS jerk is measured in rad/s$^3$. Detailed definitions are provided in Appendix~\ref{app:data-quality-metrics}.}
\label{tab:data-quality}
\centering

\resizebox{0.8\textwidth}{!}{
\begin{tabular}{@{}lcccc@{}}
\toprule
\textbf{Training Data} & \textbf{Smoothness} $\uparrow$ & \textbf{Aesthetics} $\uparrow$ & \textbf{Brightness} $-$ & \textbf{RMS jerk} $\downarrow$ \\ 
\midrule
Full human teleop & 78.1 ± 7.7 & 55.2 ± 1.6 & 133 ± 2 & 267 ± 66\\
\textsc{\method} collector & 64.2 ± 22.7 & 61.2 ± 1.1 & 130 ± 2 & 191 ± 97\\
\bottomrule
\end{tabular}}
\end{table}

Table~\ref{tab:data-quality} compares four complementary diagnostics for the two $50$-episode datasets. Smoothness combines joint-trajectory jerk with penalties for discontinuous joint transitions (higher is better), whereas Aesthetics is a heuristic image-quality score combining sharpness, colorfulness, and exposure (higher is better). Brightness checks whether illumination is comparable and has no optimization direction, while RMS jerk directly measures joint-motion variation (lower is better). Illumination is similar for full human teleoperation and \textsc{\method} ($133$ vs.\ $130$). \textsc{\method} has higher Aesthetics ($61.2$ vs.\ $55.2$) and lower RMS jerk ($191$ vs.\ $267$), but its lower Smoothness ($64.2$ vs.\ $78.1$) indicates more thresholded discontinuities and/or nonlinear episode-level effects despite the lower average RMS jerk. These diagnostics do not establish overall superiority; rather, they indicate that the collected data are comparably observable and usable while retaining a continuity gap. Full metric formulas are provided in Appendix~\ref{app:data-quality-metrics}.

\begin{table}[t]
\centering
\caption{Policy success rate after fine-tuning on different data sources for the desktop-clearing task, evaluated with 20 blind trials and no \textsc{\method} agent, VLM verifier, corrective memory, or runtime human intervention.}
\label{tab:policy}
\begin{tabular}{@{}lc@{}}
\toprule
\textbf{Data Source} & \textbf{Policy Success Rate (\%)} \\ \midrule
Scripted data & 55.0 \\
Full human teleop & 80.0 \\
\textsc{\method} data & 80.0 \\
\bottomrule
\end{tabular}
\end{table}

Table~\ref{tab:policy} provides the direct downstream evidence for data utility, with all policies trained on the same number of demonstrations. The policy fine-tuned on scripted data reaches $55.0\%$, whereas policies trained on full human teleoperation and \textsc{\method} data both reach $80.0\%$. Under this $20$-trial evaluation, the matched success rates show that \textsc{\method}-collected demonstrations can support competitive policy learning while requiring less human effort during collection (Table~\ref{tab:collection_comp}); they do not by themselves establish equivalence between the data sources. The two-dataset visual and motion diagnostics in Table~\ref{tab:data-quality} complement this downstream result by identifying comparable observability and a remaining continuity gap.

%===============================================================================

\section{Conclusion}
\label{sec:conclusion}

We presented \textsc{\method}, a human-robot symbiotic system for real-world manipulation data collection built on OpenClaw.
By reframing full autonomy as \emph{Interactive Autonomy}, the system collects, verifies, and resets autonomously behind a verification-gated loop, pausing for sparse natural-language intervention only when repeated verification failures exhaust its retry budget. 
Persistent corrections are stored in Corrective Memory and reused across subsequent rounds, while verified trajectories are consolidated to fine-tune the underlying VLA policy for deployment.

On a real-robot desktop-clearing testbed, our experiments show three main results.
First, \textsc{\method} matches the episode collection success rate of full human teleoperation while requiring substantially less human working time.
Second, the language channel repairs both verifier criteria and execution strategies, raising verifier accuracy on tested cases and improving single-attempt collection success rate across eight object categories.
Third, policies fine-tuned on \textsc{\method} data reach a policy success rate comparable to those trained on full teleoperation data, indicating that autonomously collected demonstrations can be as useful as manually collected ones when language-guided verification is in the loop.

Several limitations remain in the present report. 
Constrained by tooling and camera precision, the current system is not yet suited to autonomous collection for complex dexterous tasks.
The multi-round data flywheel described in Section~\ref{subsec:train} remains a design capability rather than a fully validated loop.
Tuning additional hyperparameters and broader task coverage are left to future work.
Ambiguous containment cases can still challenge the verifier, and object-specific execution failures may persist even after language-guided adjustments.

In the future work, we plan to extend \textsc{\method} to multi-embodiment settings and a wider range of tasks.
We view this system as a step toward scalable, human-supervised robot data collection in which operational knowledge accumulates in language rather than being discarded after every episode.

\clearpage
\setcitestyle{numbers}
\bibliographystyle{plainnat}
\bibliography{ref}
\newpage
\appendix
\section{Details of Tools used in \textsc{\method}}
\label{app:tools}

The system is packaged as four skills backed by a set of first-party pipelines. \texttt{self-learning} is a thin orchestrator that holds the plan file (\texttt{analyze\_result.yaml}); \texttt{1-analyze-task} implements the two-phase workflow of Section~\ref{subsec:collection}, generating the collection and reset plans with their judging prompts and then running the collect-and-reset loop; \texttt{understand-three-view-images} provides observation and VLM-based judging (Appendix~\ref{app:observation-tool}); and \texttt{grasp-tool} wraps the grasp-and-place pipeline and recorder (Appendices~\ref{app:grasp-tool} and~\ref{app:recorder}). Before invoking the grasp tool, the agent consults a human-readable rule file (\texttt{grasp\_experience.md}) that records verified operating rules, including arm-selection defaults, per-object approach-depth offsets, and segmentation re-prompting guidance; situations not covered by any skill follow a written out-of-distribution protocol that stops the loop, reports the observation to the operator, and waits for instructions. Hardware bindings (arm driver, camera launch, ROS topics, intrinsics, and hand-eye extrinsics) are supplied through a single configuration file, so the pipelines can be retargeted to other arms and cameras; the reference stack used in our experiments is a dual-arm Piper with Intel RealSense cameras.

\subsection{Grasp Tool}\label{app:grasp-tool}
\textbf{System overview.}
The grasp tool implements a language-conditioned pick-and-place skill used by the collection and reset routines.
Each call is specified as a quintuple $(\texttt{text\_prompt},\,\texttt{arm},\,x_{\text{place}},\,y_{\text{place}},\,\delta_{\text{depth}})$:
the text prompt selects the target, $\texttt{arm}\in\{\texttt{left},\texttt{right}\}$ chooses which Piper arm acts, $(x_{\text{place}},y_{\text{place}})$ is the release location in that arm's base frame, and $\delta_{\text{depth}}$ is an optional extra deepen along the grasp approach.
In \texttt{diligent} mode the pipeline homes both arms once, then for every object re-captures the scene, segments, plans a grasp, and executes grasp--place--home; an optional \texttt{--collect} flag records camera and joint streams to HDF5 during arm motion only.

\textbf{Perception.}
A fixed head RealSense captures aligned RGB-D ($640\times480$), which is back-projected into a colored point cloud using calibrated intrinsics and precomputed eye-to-hand extrinsics $T_{\text{cam}\rightarrow\text{base}}$ for each arm.
SAM3~\citep{sam3} segments the RGB image with the task text prompt and keeps the highest-scoring detection as a binary foreground mask.
The mask is used only for subsequent grasp filtering; the point cloud retains full scene geometry, including the table.

\textbf{Grasp planning.}
AnyGrasp~\citep{anygrasp} predicts 6-DoF grasps on the \emph{full} scene point cloud with collision detection enabled.
Candidates are then filtered in sequence:
(i) \emph{mask projection}---keep grasps whose TCP center projects into the SAM3 foreground;
(ii) non-maximum suppression and score sorting;
(iii) \emph{camera-up canonicalization}---resolve the $180^\circ$ parallel-jaw ambiguity by flipping about the approach axis so the wrist-camera axis points more upward in the arm base;
(iv) \emph{approach orientation re-ranking}---prefer oblique approaches from an elevated reference above the arm base toward the contact, blended with AnyGrasp confidence.
The top-ranked grasp is mapped from AnyGrasp TCP to the Piper flange frame via a fixed $T_{\text{TCP}\rightarrow\text{Flange}}$ and $T_{\text{base}\leftarrow\text{cam}}$, optionally shifted by the predicted grasp depth and by $\delta_{\text{depth}}$, yielding a $7$-D flange pose $(x,y,z,\text{rpy},w_{\text{grasp}})$.

\textbf{Execution.}
Pinocchio~\citep{pinocchio} IK on the Piper URDF converts waypoints to joint angles.
The arm approaches with an open gripper, closes on the object, lifts, transports to $(x_{\text{place}},y_{\text{place}})$, lowers and releases, then retracts and returns home.
Joint commands are published to \texttt{/master/joint\_\{left,right\}} at $30\,\text{Hz}$, with reach checks against puppet joint feedback.

\subsection{Observation Tool}
\label{app:observation-tool}

\textbf{Cameras and capture.}
The robot carries three RGB views: a head-mounted camera and one camera on each arm. The head camera additionally provides depth hardware-aligned to color and is the only view used for grasp planning (Appendix~\ref{app:grasp-tool}); the arm views serve observation and judging. Capture scripts subscribe to the corresponding ROS image topics, buffer the most recent frames, and save the requested views to disk; single-view (head), three-view, and RGB-D variants are provided. If the camera driver is not running, the capture wrapper starts it in a temporary session and waits up to 30 seconds for the node to come up, so an observation request never fails silently on a cold system.

\textbf{VLM querying.}
Saved views are base64-encoded and sent, together with a caller-supplied prompt, to an OpenAI-compatible endpoint serving the VLM verifier of Section~\ref{subsec:exp-setup}; a fixed preamble states that the images come from cameras on the robot and asks for a brief answer. The same interface serves three roles: free-form scene description during task analysis and planning, collection and reset judging inside the loop using the agent-authored prompts of Section~\ref{subsec:collection}, and the post-task verification hook of the collection recorder (Appendix~\ref{app:recorder}). A bounding-box variant overlays detections for prompts that require spatial grounding.

\subsection{Collection Recorder}
\label{app:recorder}

\textbf{Task interface.}
The grasp-and-place pipeline is driven by a single entry script that executes a sequence of tasks, each specified as a quintuple: a SAM3 text prompt, the executing arm, the place coordinates $(x, y)$ in that arm's base frame, and a per-task approach-depth offset in meters that deepens the grasp along the approach axis. The approach-depth offset and the arm argument are the interfaces through which the execution-side corrections of Section~\ref{subsec:exp-language} take effect. Two scheduling modes are provided: \emph{diligent} re-captures RGB-D before every task, whereas \emph{lazy} captures once and reuses the frame across tasks, which is faster but risks stale poses when an earlier task disturbs a later object. Workspace guards reject commands outside a configured safe region (place $x \in [0.05, 0.60]$\,m, $|y| \le 0.60$\,m, $|\text{offset}| \le 0.05$\,m) before any motion is sent.

\textbf{Motion phases.}
After the grasp pose is selected (Appendix~\ref{app:grasp-tool}), inverse kinematics is solved once per phase and cached, and the arm executes a fixed phase chain: home, pre-grasp posture, approach and close, lift by 0.20\,m, transport to the place coordinates, lower by 0.13\,m, release, retract by 0.08\,m, and return home. Joint targets are published at 30\,Hz with reach verification against puppet-side joint feedback.

\textbf{Episode recording.}
With collection enabled, the recorder writes the three camera streams and arm joint states to chunked HDF5, recording only while the arm is in motion; capture and model inference are never recorded. Each episode lives in its own timestamped session folder containing the episode chunks, a result file with the execution outcome and the vision verdict, the collection and verification logs, a rendered review video, and the per-episode statistics of Appendix~\ref{app:curation}. This layout realizes the per-episode logs described in Section~\ref{sec:method}.

\subsection{Data Curation and Quality Platform}
\label{app:curation}

\textbf{Per-episode statistics.}
After each episode, a background job computes action statistics (per-dimension mean, standard deviation, and range, joint path length, and gripper open and close events), the smoothness diagnostics of Appendix~\ref{app:data-quality-metrics} (RMS jerk, maximum joint jump, discontinuity count, and the 0--100 score), per-camera image quality (sharpness, brightness, contrast, exposure, and blur fraction), and the aesthetics composite. Results are written per episode, aggregated over the dataset, and rendered into a self-contained HTML report that the operator can read remotely.

\textbf{Recording-quality rating.}
A separate checker rates each episode as \emph{good}, \emph{marginal}, or \emph{bad} from its statistics. Absolute sanity rules mark an episode bad: unreadable or too-short recordings, a static arm whose joint deviations never exceed a threshold (the classic dead-topic failure in which the recorder ran but the robot never moved), a dark camera, or an excessive number of joint discontinuities. Dataset-relative rules mark an episode marginal when its RMS jerk or per-camera sharpness is an outlier against the dataset median. The rating is independent of task success, as stated in Section~\ref{subsec:labeling}.

\textbf{Offline re-judging.}
Because each episode retains its final observations, a dedicated tool re-judges task success offline from the last recorded frames of an episode, querying the same endpoint as the live verifier with an overridable prompt. This is the mechanism that lets a corrected judging criterion be applied retroactively, so the whole dataset stays labeled under one standard (Appendix~\ref{app:data}).

\textbf{Training-set preparation.}
A consolidation tool selects qualified sessions, requiring script-level completion, a passing vision verdict, and optionally the recording-quality filter, merges their chunked recordings into contiguous training episodes, and maintains a manifest with a frozen ordering so that every training set in Section~\ref{sec:exp} can be reproduced exactly.

\section{Training and Evaluation Protocols}
\label{app:train-eval}

This appendix details the fine-tuning and deployment evaluation procedures referred to in Section~\ref{subsec:train}. Both follow standard practice and are included for completeness and reproducibility rather than as contributions of this work.

\subsection{VLM Verifier Prompts}
\label{app:verifier-prompts}

At the end of each collection or reset phase, the loop captures a fresh camera image and queries Seed1.8 with a phase-specific judging prompt produced during task analysis.
The verifier returns a binary success/failure decision: the prompt asks the model to answer \texttt{YES} or \texttt{NO} first, and the loop treats \texttt{YES} as a pass and \texttt{NO} as a fail that triggers retry (or an \textsc{Alert} once retries are exhausted).
The prompts below follow the desktop-clearing plan in which target objects are placed into a basket and later returned to the table; the agent may generate task-specific variants at plan time, and persistent corrections from Corrective Memory edit the relevant fields.

\noindent\textbf{Collection success criterion.}
Target objects are inside the basket.

\noindent\textbf{Collection judging prompt.}
\texttt{Are the target objects inside the basket? Answer YES or NO first.}

\noindent\textbf{Reset success criterion.}
Target objects are back on the table outside the basket.

\noindent\textbf{Reset judging prompt.}
\texttt{Are the target objects back on the table (not in the basket)? Answer YES or NO first.}

\subsection{Data Consolidation}
\label{app:data}

A run produces a log in which every trajectory carries the verifier's verdict and the context under which it was collected. Consolidation proceeds in three steps.

\textbf{Filtering.} Each trajectory is kept or discarded by the collection success criterion in the state the operator left it at the end of the run. Trajectories the verifier marked as failures, and trajectories collected before a criterion was corrected and re-judged as failures under the corrected criterion, are dropped. Concretely, when a prompt rule is added to Corrective Memory we re-evaluate the affected trajectories against the updated criterion so that the entire dataset is labeled under one consistent standard rather than the standard that happened to be in force when each trajectory was recorded.

\begin{figure}[H]
\centering
\includegraphics[width=\linewidth]{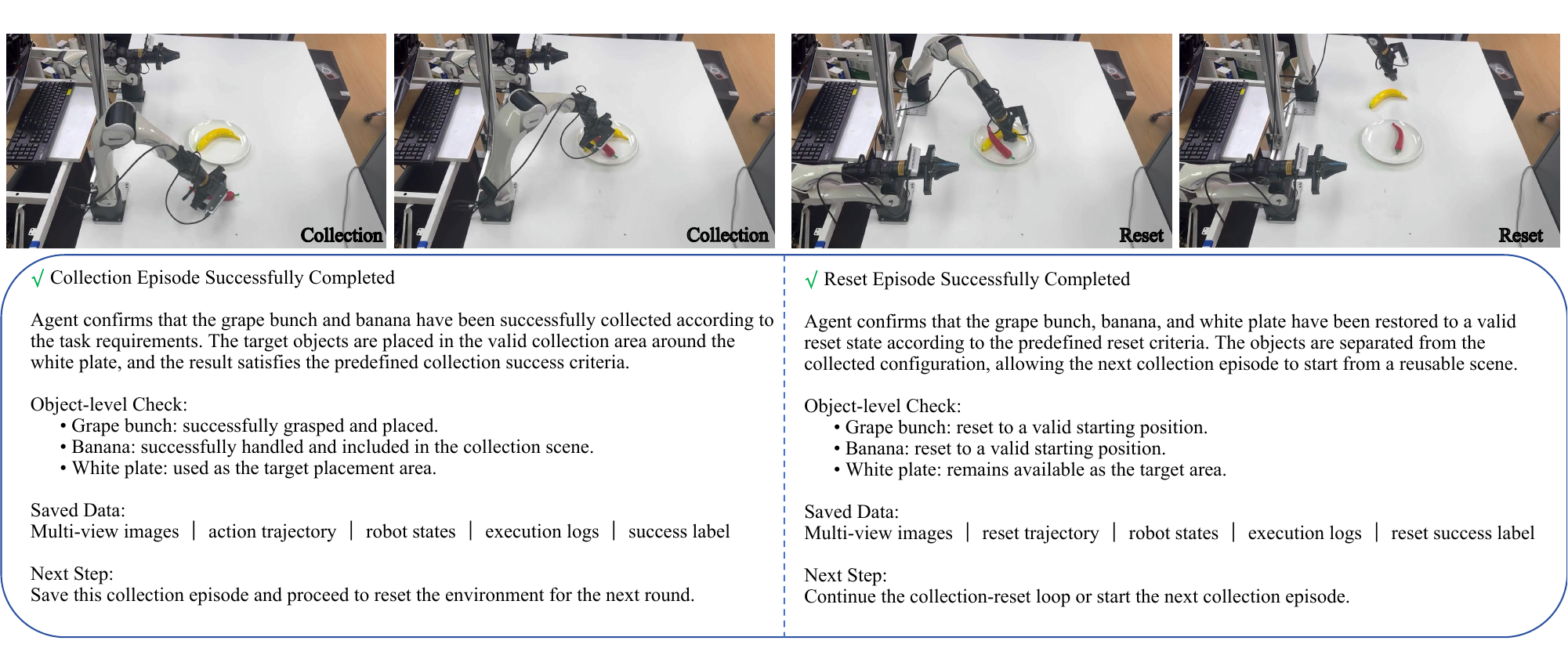}
\caption{Representative feedback for successful collection-reset episodes. The system checks whether the grape bunch, banana, and white plate satisfy the predefined collection or reset criteria. Successful episodes proceed to data saving and the next step.}
\label{fig:app-collection-reset-feedback}
\end{figure}

Figure~\ref{fig:app-collection-reset-feedback} shows representative success feedback used during collection and reset verification. These examples illustrate how successful episodes are confirmed and retained before data consolidation.

\textbf{Quality report.} Over the surviving trajectories we compute summary statistics of the recorded actions, the per-dimension mean and standard deviation of the action vectors and a diversity measure given by the mean pairwise distance between trajectory action sequences. These are reported as a read-only description of the dataset. They are not used to accept or reject individual trajectories and they do not trigger any re-collection; a low diversity reading is reported, not acted on. This is the deliberate choice noted in Section~\ref{subsec:train} to leave coverage-driven re-collection to future work.

\textbf{Packaging.} The kept trajectories are written in the observation-action format expected by the policy's fine-tuning interface, with each record pairing the camera observation at a step with the action taken at that step.

\subsection{Fine-Tuning}
\label{app:finetune}

We fine-tune $\pi_{0.5}$~\citep{pi05} on each consolidated dataset, following the training setting of~\citet{pi05}.
To keep comparisons across collection conditions meaningful, this configuration is held fixed across datasets: the same base checkpoint, optimizer, schedule, and number of update steps.

\subsection{Deployment Evaluation}
\label{app:eval}

Each fine-tuned policy is tested on the physical platform with no agent orchestrating it and no VLM verifying its actions: the policy receives a camera observation and outputs actions directly, exactly as it would in deployment. This isolates the quality of the learned policy from the collection-time machinery, as argued in Section~\ref{subsec:train}.

\textbf{Protocol.} We run 20 blind deployment trials on the desktop-clearing task from randomized initial placements. A trial counts as a success when the policy completes the task within the criterion used during collection. We report the overall policy success rate across these trials. Initial placements are drawn from the same distribution across the policies being compared, so that differences in success rate reflect the policies and not the test conditions.

\textbf{Use in the experiments.} Section~\ref{sec:exp} compares policies fine-tuned on datasets from different collection conditions under this single protocol. Because filtering, fine-tuning, and the test protocol are all held fixed across conditions, a difference in policy success rate between two policies is attributable to the data each was trained on.

\section{Experiment Details}
\label{app:exp}

This appendix collects supplementary material for the experiments in Section~\ref{sec:exp}: representative scenes for the verifier-correction settings of Table~\ref{tab:language_correction} (Appendix~\ref{app:verifier-scenes}), definitions of the data-quality metrics reported in Table~\ref{tab:data-quality} (Appendix~\ref{app:data-quality-metrics}), and a per-episode distribution comparison of those metrics for the human and agent datasets (Appendix~\ref{app:dataset-metrics-comparison}).

\subsection{Verifier-Correction Scenes}
\label{app:verifier-scenes}
 
\begin{figure}[t]
\centering
\includegraphics[width=1.00\linewidth]{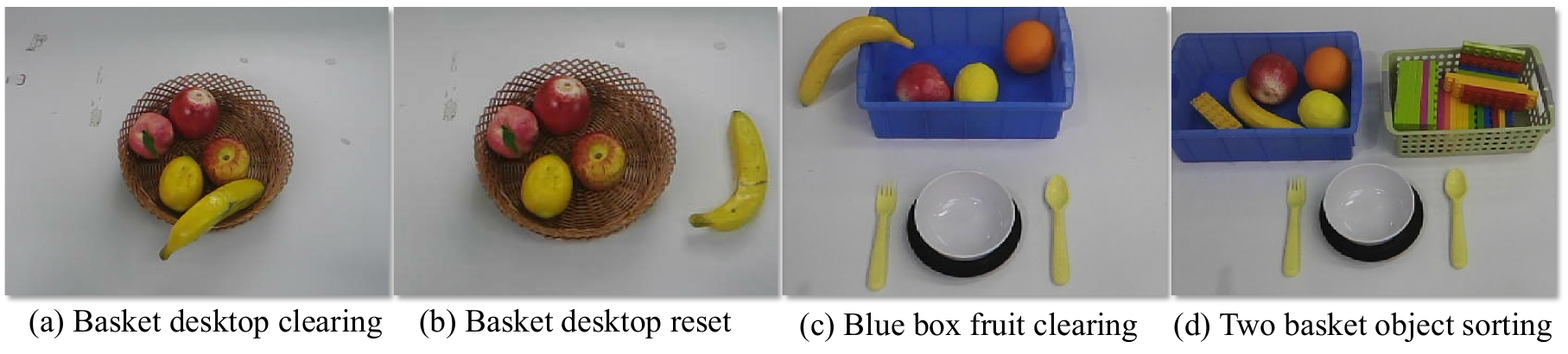}
\caption{Representative scenes for the four verifier-correction settings evaluated in Table~\ref{tab:language_correction}: (a) basket desktop clearing, (b) basket desktop reset, (c) blue-box fruit clearing, and (d) two-basket object sorting. }
\label{fig:basket-verifier-cases}
\end{figure}

Table~\ref{tab:language_correction} evaluates the language-corrected judging rules across four settings that differ in container and layout. Figure~\ref{fig:basket-verifier-cases} shows a representative scene for each. Panels (a) and (b) depict the basket desktop-clearing task on which the criterion mismatch was identified: the collection stage (a) is judged under the partial-containment rule, and the reset stage (b) under the relaxed reset criterion. Panels (c) and (d) show the two layouts used to test whether the corrected rules transfer: blue-box fruit clearing (c), the harder setting in which containment judgments are frequently ambiguous, and two-basket object sorting (d), in which the judgment additionally conditions on object category. Each row of Table~\ref{tab:language_correction} reports accuracy over ten judgment cases drawn from scenes of the corresponding setting.
 
\subsection{Data-Quality Metrics}
\label{app:data-quality-metrics}

\noindent\textbf{Scope and aggregation. }
Table~\ref{tab:data-quality} summarizes two datasets, each containing $N=50$ episodes. For each metric, the implementation first computes one value $x_i$ per episode and then reports the unweighted episode mean and sample standard deviation,
\begin{equation}
    \bar{x}=\frac{1}{N}\sum_{i=1}^{N}x_i,
    \qquad
    s=\sqrt{\frac{1}{N-1}\sum_{i=1}^{N}(x_i-\bar{x})^2}.
\end{equation}
Thus, the $\pm$ notation denotes sample standard deviation rather than standard error. Image-based episode values are averaged hierarchically over sampled frames, then cameras, and finally episodes. Below, $\operatorname{clip}(z,a,b)=\min\{\max\{z,a\},b\}$.

\noindent\textbf{Joint-signal metrics.}
Let $q_{t,k}$ denote the position of arm joint $k$ at frame $t$. We use the 12 arm joints (six per arm), exclude both grippers, and sample at 30~Hz, so $\Delta t=1/30$~s. The third forward finite-difference jerk is
\begin{equation}
    j_{t,k}
    =\frac{q_{t+3,k}-3q_{t+2,k}+3q_{t+1,k}-q_{t,k}}
    {\Delta t^3}.
\end{equation}
For an episode of $T$ frames, the reported RMS jerk is
\begin{equation}
    J_{\mathrm{RMS}}
    =\sqrt{\frac{1}{12(T-3)}
    \sum_{t=0}^{T-4}\sum_{k=1}^{12}j_{t,k}^{2}},
\end{equation}
with units of $\mathrm{rad}/\mathrm{s}^{3}$; lower values indicate smoother joint motion. The implementation-specific \emph{Smoothness} diagnostic additionally counts thresholded frame-to-frame discontinuities,
\begin{equation}
    D=\sum_{t=0}^{T-2}
    \mathbf{1}\!\left\{
    \max_{1\leq k\leq 12}|q_{t+1,k}-q_{t,k}|>0.10~\mathrm{rad}
    \right\},
\end{equation}
then computes
\begin{equation}
    S=\operatorname{clip}\!\left(
    100\exp\!\left(-\frac{J_{\mathrm{RMS}}}{1200}\right)-5D,
    0,100\right).
\end{equation}
Higher $S$ is better. Because each detected discontinuity incurs a five-point penalty, a dataset can have lower RMS jerk yet lower Smoothness if its episodes contain more thresholded discontinuities.

\noindent\textbf{Image metrics.}
For a grayscale sampled frame $Y_f\in[0,255]^{H\times W}$, brightness is
\begin{equation}
    B_f=\frac{1}{HW}\sum_{u=1}^{H}\sum_{v=1}^{W}Y_f(u,v).
\end{equation}
Brightness ranges from 0 to 255 and has no optimization direction; it is reported as an illumination-comparability check.

The implementation-specific \emph{Aesthetics} diagnostic is the custom heuristic
\begin{equation}
    A_f=100\left(0.4s_f+0.3c_f+0.3e_f\right).
\end{equation}
Its sharpness component is
\begin{equation}
    s_f=\operatorname{clip}\!\left(
    \frac{\log_{10}\!\left(1+\operatorname{Var}(\operatorname{Laplacian}(Y_f))\right)}{3},
    0,1\right).
\end{equation}
For the Hasler--Suesstrunk-style colorfulness component, let
\begin{equation}
    r_g=R-G,
    \qquad
    y_b=\frac{R+G}{2}-B,
\end{equation}
where $\mu_{r_g},\mu_{y_b}$ and $\sigma_{r_g},\sigma_{y_b}$ are their pixelwise means and standard deviations. Then
\begin{equation}
    C_f=\sqrt{\sigma_{r_g}^{2}+\sigma_{y_b}^{2}}
    +0.3\sqrt{\mu_{r_g}^{2}+\mu_{y_b}^{2}},
    \qquad
    c_f=\operatorname{clip}\!\left(\frac{C_f}{80},0,1\right).
\end{equation}
Defining the saturated-pixel fractions as
\begin{equation}
    p_{\mathrm{over}}=\frac{\bigl|\{(u,v):Y_f(u,v)\geq250\}\bigr|}{HW},
    \qquad
    p_{\mathrm{under}}=\frac{\bigl|\{(u,v):Y_f(u,v)\leq5\}\bigr|}{HW},
\end{equation}
the exposure component is
\begin{equation}
    e_f=\exp\!\left[-\left(\frac{B_f-128}{64}\right)^2\right]
    \operatorname{clip}\!\left(1-4(p_{\mathrm{over}}+p_{\mathrm{under}}),0,1\right).
\end{equation}
The resulting Aesthetics score ranges from 0 to 100, with higher values preferred.

\subsection{Dataset Metrics Comparison}
\label{app:dataset-metrics-comparison}

\begin{figure}[H]
\centering
\includegraphics[width=\linewidth]{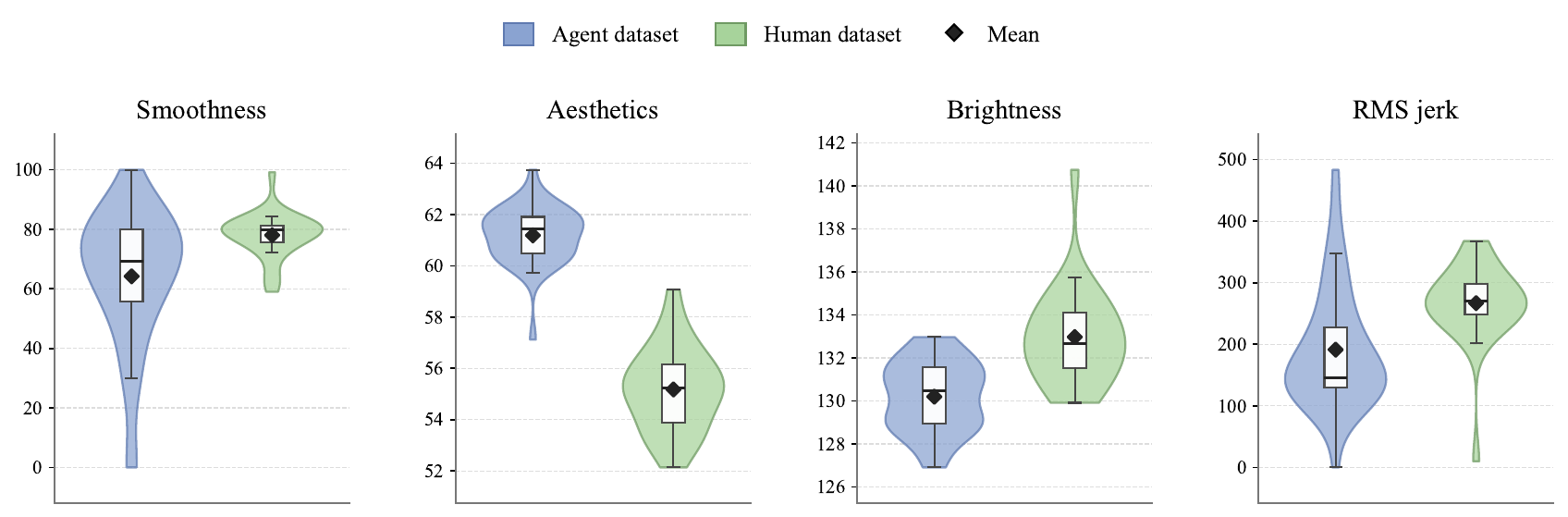}
\caption{Per-episode distributions of Smoothness, Aesthetics, Brightness, and RMS jerk for the agent-collected dataset and the human teleoperation dataset ($N=50$ episodes each). Diamonds mark episode means; boxes show quartiles and medians.}
\label{fig:dataset-metrics-comparison}
\end{figure}

Figure~\ref{fig:dataset-metrics-comparison} shows the per-episode distributions of the four diagnostics in Table~\ref{tab:data-quality} for the two $50$-episode datasets of Section~\ref{subsec:exp-quality}: the human teleoperation set and the \textsc{\method} (agent) set.
Each panel is a violin-and-boxplot for one metric---Smoothness, Aesthetics, Brightness, and RMS jerk---with the episode mean marked by a diamond.
The spreads match the aggregate trends in Table~\ref{tab:data-quality}: brightness is comparable; the agent set has higher Aesthetics and lower mean RMS jerk; and its Smoothness is lower on average with a wider spread, consistent with more thresholded discontinuities among agent-collected episodes.

\end{document}